\documentclass[conference]{IEEEtran}
\IEEEoverridecommandlockouts

\usepackage{cite}
\usepackage{amsmath,amssymb,amsfonts}
\usepackage{algorithmic}
\usepackage{algorithm}
\usepackage{array}
\usepackage{graphicx}
\usepackage{textcomp}
\usepackage{xcolor}
\usepackage{booktabs}
\usepackage{multirow}
\usepackage{url}
\usepackage{enumitem}
\usepackage{stfloats}
\usepackage{placeins}   

\begin{document}

\title{GameUIAgent: An LLM-Powered Framework for Automated Game UI
Design with Structured Intermediate Representation}

\author{Wei~Zeng, Fengwei~An, Zhen~Liu, and~Jian~Zhao%
\thanks{This work is supported by the Zhongguancun Academy (Grant No. C20250508). (Corresponding author: Jian Zhao.)}%
\thanks{W. Zeng and F. An are with the School of Microelectronics, Southern University of Science and Technology (SUSTech), Shenzhen, China.}%
\thanks{Z. Liu is with the School of Marxism, Tsinghua University, Beijing, China.}%
\thanks{W. Zeng, X. Li, Z. Liu, and J. Zhao are with the Zhongguancun Academy, Beijing, China (e-mail: jianzhao@zgci.ac.cn).}%
}

\maketitle

\begin{abstract}
Game UI design requires consistent visual assets across rarity tiers
yet remains a predominantly manual process. We present
\textbf{GameUIAgent}, an LLM-powered agentic framework that translates
natural language descriptions into editable Figma designs via a
\textit{Design Spec JSON} intermediate representation. A six-stage
neuro-symbolic pipeline combines LLM generation, deterministic
post-processing, and a Vision-Language Model (VLM)-guided \textit{Reflection Controller} (RC)
for iterative self-correction with guaranteed non-regressive quality.
Evaluated across 110 test cases, three LLMs, and three UI templates,
cross-model analysis establishes a \textbf{game-domain failure taxonomy}
(\textit{rarity-dependent degradation}; \textit{visual emptiness}) and
uncovers two key empirical findings. A \textbf{Quality Ceiling Effect}
(Pearson $r\!=\!-0.96$, $p\!<\!0.01$) suggests that RC improvement is
bounded by headroom below a quality threshold---a visual-domain
counterpart to test-time compute scaling laws. A
\textbf{Rendering-Evaluation Fidelity Principle} reveals that partial
rendering enhancements paradoxically degrade VLM evaluation by
amplifying structural defects. Together, these results establish
foundational principles for LLM-driven visual generation agents in
game production.
\end{abstract}


\section{Introduction}

Game user interface (UI) design fundamentally shapes player
experience~\cite{fullerton2019game, schell2020art}. Visual assets such
as character cards and inventory thumbnails must balance aesthetic
appeal, functional clarity, and thematic coherence---often across
hundreds of elements spanning hierarchical rarity tiers (e.g.,
N/R/SR/SSR/UR in gacha-style games where collectible rarity governs
visual fidelity). This remains a predominantly manual,
labor-intensive process that constitutes a significant bottleneck in
game art production pipelines~\cite{adams2014fundamentals}.

Existing generative AI solutions typically target web interfaces or
produce static raster images lacking editability. Figma-integrated
approaches---Figma Make~\cite{figma2025make}, GUIDE~\cite{guide2025},
and CoGen~\cite{cogen2026}---operate single-shot without quality
verification and lack game-domain rules such as rarity hierarchies. We
propose \textbf{GameUIAgent}, a neuro-symbolic pipeline that decouples
LLM creative generation from deterministic rendering via a \textit{Design
Spec JSON} intermediate representation, closing the perception-action
loop through an independent Vision-Language Model (VLM)-guided \textit{Reflection Controller}
(RC).

An experimental finding motivates a key design constraint: adding
gradient rendering \emph{without} correcting layout intent
\emph{worsens} VLM evaluation, because enhanced rendering exposes
structural defects previously concealed by flat colors. We term this
the \textbf{Rendering-Evaluation Fidelity Principle}. This paper makes
three primary contributions:

\begin{enumerate}[noitemsep, topsep=2pt, leftmargin=*]
\item \textbf{GameUIAgent and Design Spec JSON}: an end-to-end pipeline
  from natural language to editable Figma game UI designs with
  guaranteed non-regressive quality improvement.
\item A \textbf{game-domain failure taxonomy} for LLM-based structured
  UI generation: two task-structural failure modes---\textit{rarity-dependent
  degradation} (complexity-overload in Gemini; inverse over-generation in
  GPT-4o-mini across the N/R/SR/SSR/UR hierarchy) and \textit{visual
  emptiness} (syntactically valid JSON yielding blank renders)---establish
  that JSON validity is necessary but insufficient for design quality.
  Five objective structural metrics further reveal a \textit{perceptual
  blind spot}: few-shot scaffolding governs compositional richness
  ($+38\%$ Node Count, $+42\%$ Color Diversity) invisible to VLM
  perceptual evaluation yet critical for production-grade output.
\item \textbf{Two empirical findings}: the \textbf{Quality Ceiling Effect}
  bounding iterative self-correction (Pearson $r\!=\!-0.96$ with initial
  quality), and the \textbf{Rendering-Evaluation Fidelity Principle}
  demonstrating how mismatched rendering fidelity can invert VLM reward
  signals---establishing foundational design principles for LLM-driven
  visual generation agents.
\end{enumerate}

\section{Related Work}

\textbf{AI in Game UI Design.}
Recent surveys~\cite{gallotta2024llm} identify UI asset creation as a
key emerging LLM application in games. pix2code~\cite{beltramelli2018pix2code}
and Design2Code~\cite{si2024design2code} pioneered UI-to-code generation
for web interfaces, while LayoutGPT~\cite{feng2024layoutgpt} demonstrates
compositional layout planning. These methods target standard interfaces
ill-suited to the non-standard layouts of game UI. Two concurrent works
address structured game UI directly: AutoGameUI~\cite{autogameui2024}
assembles UIs from existing designs via multimodal matching; SpecifyUI~\cite{specifyui2025}
extracts hierarchical IR from reference screenshots using designer
feedback. Both differ fundamentally from GameUIAgent, which performs
text-to-new-design generation combined with automated quality
verification.

\textbf{LLMs for Iterative Refinement and Evaluation.}
ReAct~\cite{yao2022react}, Reflexion~\cite{shinn2023reflexion}, and
SELF-REFINE~\cite{madaan2023self} demonstrate LLM improvement via verbal
feedback. A critical limitation remains: LLMs cannot reliably self-correct
when acting as their own judges~\cite{huang2024cannot, kamoi2024correct},
and self-improvement can degrade performance~\cite{zhao2026faithful}.
GameUIAgent mitigates this failure mode by sourcing feedback from an
\emph{independent} VLM critic and employing state-tracking to guarantee
non-regressive improvement. The LLM-as-Judge paradigm~\cite{zheng2023judging}
underpins our evaluator; to our knowledge, we provide the first
Intraclass Correlation Coefficient [ICC(2,1)]~\cite{shrout1979intraclass, cicchetti1994guidelines}
reliability quantification for VLM-as-Judge on visual design quality.

\textbf{Test-Time Compute Scaling.}
Recent work demonstrates that extending inference compute via iterative
refinement can outperform scaling model parameters in certain
regimes~\cite{snell2024scaling}. A critical open question is \emph{when}
additional refinement cycles yield diminishing returns. Our Quality
Ceiling Effect provides a principled answer in the visual design domain:
gain is bounded by evaluator headroom, not generator capacity, once
output quality approaches the evaluator's distinguishability
threshold---making evaluator quality the primary lever for scaling
agentic iterative refinement.

\section{System Design}

\subsection{Overview}

\begin{figure*}[!tb]
\centering
\includegraphics[width=\textwidth]{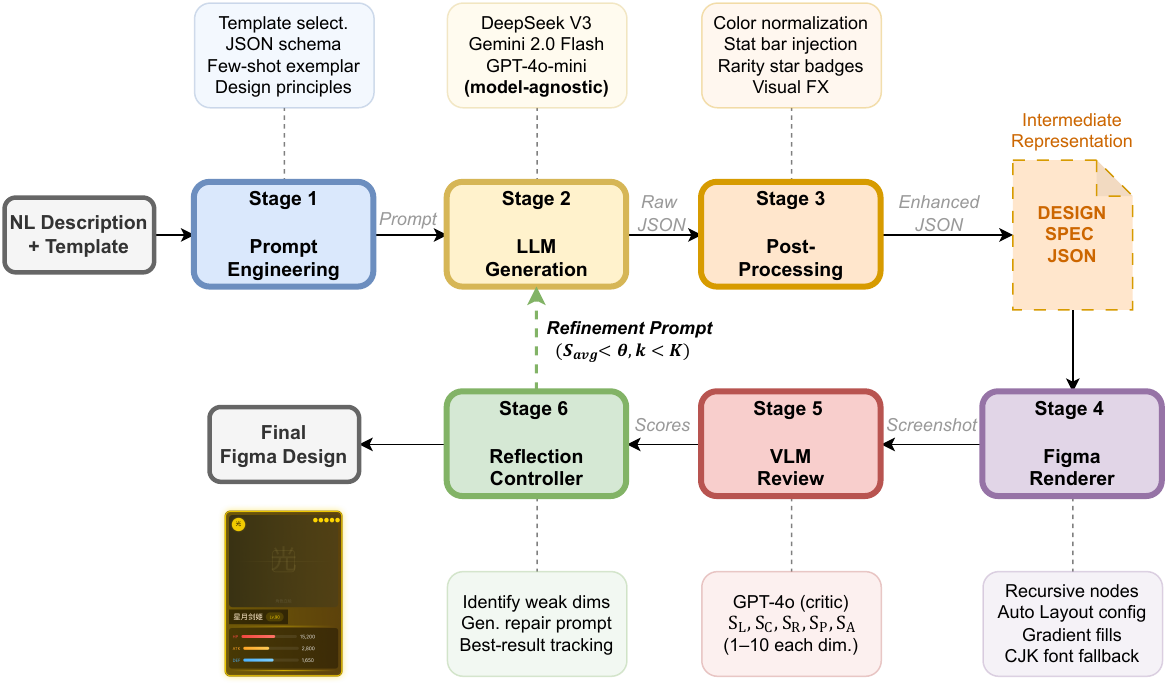}
\caption{Overall architecture of GameUIAgent. Stages~1--5 constitute the
forward generation pass; Stage~6 (Reflection Controller, green) closes
the agentic loop by converting VLM quality scores into targeted repair
prompts and re-invoking Stage~2 if $S_{\text{avg}}\!<\!\theta$. The
\textit{Design Spec JSON} intermediate representation (dashed box)
separates creative generation from deterministic enhancement.}
\label{fig:architecture}
\end{figure*}

Fig.~\ref{fig:architecture} illustrates GameUIAgent's six-stage pipeline
bridging natural language intent with professional design tools.
Consistent with established agent frameworks~\cite{wang2024survey,
anthropic2025guide}, the architecture integrates LLM-based creative
generation, deterministic post-processing, and a self-correcting
VLM-driven Reflection Controller. Concretely, the six stages are:
\textbf{(1)} prompt engineering,
\textbf{(2)} LLM generation of Design Spec JSON,
\textbf{(3)} intelligent post-processing (repair, data injection, rarity enhancement),
\textbf{(4)} renderer-based rasterization (Figma plugin or Python preview),
\textbf{(5)} VLM quality review, and
\textbf{(6)} Reflection Controller agentic loop. Three templates are supported:
\textit{Character Card} (320$\times$450\,px, vertical hierarchy),
\textit{Item Thumbnail} (96$\times$96\,px, compact grid), and
\textit{Skill Panel} (280$\times$360\,px, list composition). All prompts
are centralized in a unified module; the system is model-agnostic,
supporting any OpenAI-compatible LLM. A Python backend exposes HTTP REST endpoints that the Figma plugin
queries to obtain finalized Design Spec JSON; a lightweight Python
preview renderer enables high-throughput VLM evaluation without
interactive Figma sessions.

\subsection{Design Spec JSON}
\label{sec:design_spec}

The \textit{Design Spec JSON} schema is the central enabling
contribution. It is \textit{LLM-friendly} (natural naming aligned with
pre-training data distributions), \textit{tool-agnostic} (grounded in
universal vector-graphics primitives transferable to Sketch or Adobe XD),
and \textit{expressive} (encoding gradient fills, drop shadows, and
auto-layout constraints). The schema defines a recursive node tree in
which each node specifies geometry (\texttt{x},\,\texttt{y},\,\texttt{width},\,\texttt{height}),
visual style (\texttt{fills},\,\texttt{strokes},\,\texttt{effects}), text
content (\texttt{characters}), and hierarchical relationships
(\texttt{children}). Four node types are supported: FRAME, RECTANGLE,
ELLIPSE, and TEXT. A token budget of 6,000 per call accommodates complex
UR-tier designs (average 4,534 tokens for DeepSeek V3).

\subsection{Intelligent Post-Processing}

Raw LLM outputs are enhanced in three deterministic phases.
\textbf{(1) Repair}: color values are normalized (0--255 integer RGB to
0--1 float) and geometric dimensions clamped to positive integers.
\textbf{(2) Data Injection}: stat bar widths are computed algorithmically
from attribute values, achieving numerical precision beyond standard LLM
capabilities.
\textbf{(3) Rarity Enhancement}: a \textit{Rarity Progression System}
injects tier-scaled visual decorators across N/R/SR/SSR/UR---including
star badges (0--5 icons), progressively upgraded border styles (simple
strokes at N through multi-layer glow at UR), and element-specific
thematic gradients.

\subsection{VLM-Based Quality Review}
\label{sec:review}

A VLM evaluates generated designs across five dimensions:
\textit{Layout} ($S_L$, alignment and spacing), \textit{Consistency}
($S_C$, color themes and font hierarchy), \textit{Readability} ($S_R$,
text contrast), \textit{Completeness} ($S_P$, presence of required game
elements), and \textit{Aesthetics} ($S_A$, visual appeal and rarity
appropriateness). Each dimension is scored 1--10 under a standardized
rubric applied uniformly across all evaluation contexts.

\subsection{Reflection Controller}
\label{sec:reflection}

The Reflection Controller (Algorithm~\ref{alg:reflection}) implements a
self-correcting agentic loop inspired by SELF-REFINE~\cite{madaan2023self}
and Reflexion~\cite{shinn2023reflexion}. Given VLM scores
$\mathbf{s} = (S_L, S_C, S_R, S_P, S_A)$, the controller identifies the
two lowest-scoring dimensions and formulates a targeted Refinement Prompt
with dimension-specific repair instructions (e.g., ``Increase text
contrast ratio $>$ 4.5:1'' for low $S_R$; ``Add missing
\texttt{rarity\_stars} nodes'' for low $S_P$). This \emph{surgical}
repair strategy preserves correct elements while correcting only flagged
deficiencies. The loop terminates when $S_{\text{avg}} \geq \theta$
(convergence) or $k = K$ (budget exhaustion), performing up to $K\!+\!1$
VLM evaluations in total ($k = 0, 1, \ldots, K$); best-result tracking
guarantees the returned Design Spec is never inferior to initial output.
The sampling temperature is reduced from 0.7 to 0.5 during refinement to
encourage conservative corrections.

\begin{algorithm}[!ht]
\caption{Reflection Controller with Best-Result Tracking}
\label{alg:reflection}
\begin{algorithmic}
\STATE \textbf{Input:} $\text{spec}_0$,\ $\theta$ (threshold),\ $K$ (max iter)
\STATE \textbf{Output:} $\text{spec}^*$ (best Design Spec)
\STATE $\text{spec}^* \leftarrow \text{spec}_0$,\ $s^* \leftarrow 0$,\ $k \leftarrow 0$
\WHILE{$k \leq K$}
    \STATE $I_k \leftarrow \textsc{Render}(\text{spec}_k)$
    \STATE $\mathbf{s}_k \leftarrow \textsc{VLMReview}(I_k,\,\text{rubric})$
    \IF{$s_{\text{avg},k} > s^*$}
        \STATE $\text{spec}^* \leftarrow \text{spec}_k$,\ $s^* \leftarrow s_{\text{avg},k}$
    \ENDIF
    \IF{$s_{\text{avg},k} \geq \theta$ \OR $k = K$}
        \STATE \textbf{break}
    \ENDIF
    \STATE $\mathcal{W} \leftarrow \textsc{Bot2}(\mathbf{s}_k)$ \COMMENT{Two lowest dims}
    \STATE $p_{\text{refine}} \leftarrow \textsc{GenRepairPrompt}(\mathcal{W},\,\mathbf{s}_k)$
    \STATE $\text{spec}_{k+1} \leftarrow \textsc{LLMRefine}(\text{spec}_k,\,p_{\text{refine}},\,T\!=\!0.5)$
    \STATE $k \leftarrow k + 1$
\ENDWHILE
\STATE \textbf{return} $\text{spec}^*$
\end{algorithmic}
\end{algorithm}

The controller constitutes a finite-horizon Markov Decision Process (MDP) where defining
the episode return as $G = \max_{0 \leq t \leq K} S_{\text{avg}}(\hat{q}_t)$
ensures best-result tracking yields non-regressive quality, precluding the
score regression common in self-refining agents.

\section{Experiments}

Our experiments address four research questions.
\textbf{RQ1 (Failure Taxonomy)}: what failure modes and reliability
dimensions govern LLM-based structured game UI generation across models
with different optimization targets.
\textbf{RQ2 (Component Contribution)}: impact of individual prompt
components and post-processing.
\textbf{RQ3 (Generalization)}: adaptability across UI templates and
renderer configurations.
\textbf{RQ4 (Agentic Refinement)}: RC effectiveness and the Quality
Ceiling Effect. Code and data available at https://github.com/zengwei-code/GameUIAgent.

\begin{figure*}[!t]
\centering
\includegraphics[width=\textwidth]{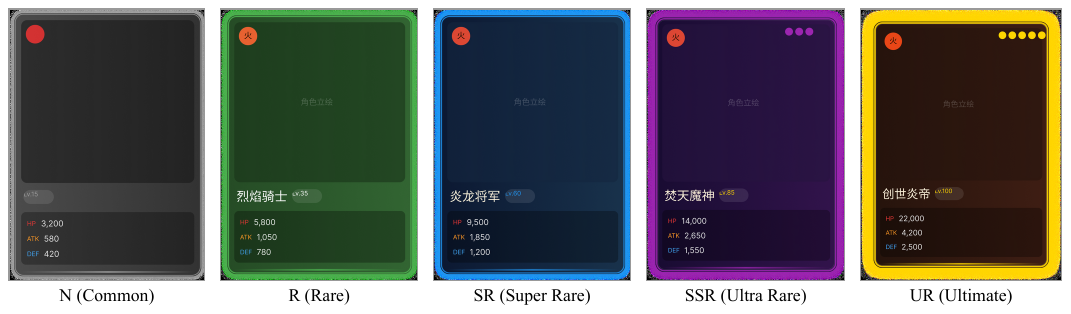}
\caption{Rarity progression (N to UR) generated by GameUIAgent with
DeepSeek V3. Progressively richer visual treatments are injected
automatically: simple gray borders (N) $\to$ colored theme borders (R)
$\to$ gradient fills with glow (SR) $\to$ multi-layered borders with
star badges (SSR) $\to$ golden ornate frames (UR). All five cards share
the Fire element theme, demonstrating consistent thematic adaptation
across the rarity hierarchy.}
\label{fig:rarity}
\end{figure*}

Fig.~\ref{fig:rarity} presents a qualitative overview of GameUIAgent's
outputs across all five rarity tiers (DeepSeek V3, Fire element theme),
illustrating the progressive visual enrichment injected automatically by
the post-processing pipeline. Quantitative evaluation of these outputs
follows in the sections below.

\subsection{Experimental Setup}
\label{sec:experimental_setup}

\textbf{Models.} The framework accepts any OpenAI-compatible API endpoint,
enabling straightforward replication with current-generation models. Three
representative LLMs spanning distinct optimization targets are evaluated via
OpenRouter to characterize failure modes across the quality--cost--speed
design space:
\textbf{DeepSeek V3}~\cite{deepseekai2024deepseek} (MoE, 671B params,
quality-optimized, \$0.008/call) serves as the primary generator;
\textbf{Gemini 2.0 Flash}~\cite{google2025gemini2flash} represents
speed- and cost-optimized inference (\$0.001/call);
\textbf{GPT-4o-mini}~\cite{openai2024gpt4omini} represents
general-purpose instruction-following at moderate scale (\$0.012/call).
\textbf{GPT-4o}~\cite{openai2024gpt4o} serves as the automated VLM critic.

\textbf{Templates and Test Cases.} 110 cases total: \textit{Character
Card} (50 cases: 5 rarity tiers $\times$ 10 characters per tier,
spanning 5 element types: Fire, Water, Wind, Light, Dark),
\textit{Item Thumbnail} (30 cases, 96$\times$96\,px), and
\textit{Skill Panel} (30 cases, list-based composition).

\textbf{Metrics.} JSON Validity Rate ($V$), VLM Quality Scores ($S$,
1--10 per dimension), latency ($T$), and token count. Three progressively
capable renderer configurations enable controlled comparisons: \textit{Flat}
(solid colors only), \textit{Gradient} (adds linear fills and drop
shadows), and \textit{Layout-aware} (adds Auto Layout simulation resolving
child coordinate conflicts prior to rasterization). Five structural
metrics---Node Count (NC), Canvas Coverage (COV), Color Diversity (CD),
Text Contrast Ratio (CR), Element Completeness (EC)---are computed
directly from Design Spec JSON without rendering.

\textbf{Evaluator Reliability.} Three independent GPT-4o runs on 49
DeepSeek V3 designs yield within-VLM ICC(2,1)$\,=\,0.555$~\cite{shrout1979intraclass,
cicchetti1994guidelines} (fair, per Cicchetti's guidelines), with stable aggregate means
(8.02, 8.03, 8.06; mean per-design MAD$\,=\,0.15$). Since the RC compares successive
iterations of the \emph{same} design under the \emph{same} evaluator,
within-VLM reliability is the operative metric validating single-evaluator
RC design. A supplementary multi-VLM panel (GPT-4o~\cite{openai2024gpt4o},
Claude-3.5-Sonnet~\cite{anthropic2024claude35},
Gemini~2.0~Flash~\cite{google2025gemini2flash}; $n\!=\!20$ stratified designs) reveals \textbf{VLM
Quality Perception Divergence}: cross-VLM ICC(2,1)$\,=\,0.021$,
K$\alpha\,=\,-0.088$~\cite{krippendorff2004}, driven by scale saturation
(GPT-4o $\sigma\,=\,0.40$ vs.\ Gemini $\sigma\,=\,0.28$) and substantive
disagreement on Completeness (GPT-4o: 5.65 vs.\ Claude: 7.15, reflecting
differing views on Figma-native vs.\ schema-level coverage). Aesthetics
is the sole dimension with positive cross-VLM agreement ($\alpha\,=\,0.023$),
confirming holistic visual appeal as the most model-invariant dimension.

\subsection{RQ1: LLM Failure Taxonomy and Cross-Model Analysis}
\label{sec:rq1}

Table~\ref{tab:model_comparison} reveals that \textit{structural
reliability} (JSON validity) and \textit{per-output perceptual quality}
are orthogonal performance dimensions in structured UI generation:
Gemini~2.0~Flash (88\%, 6.4/10) and GPT-4o-mini (56\%, 6.7/10) achieve
near-equivalent VLM scores despite a 32-point validity gap, confirming
that schema conformance reliability---not single-output perceptual
quality---is the primary cross-model differentiator. All pairwise quality
differences are statistically significant (Mann-Whitney $p\!<\!0.001$).
DeepSeek V3 leads on both dimensions (98.0\% validity, 8.0/10,
$S_C\!=\!8.9$), establishing it as the primary generator for RQ2--RQ4.
As noted in Table~\ref{tab:model_comparison}, DeepSeek V3 scores reflect
Figma renders while others use preview renders; the renderer comparison
(\S\ref{sec:rq3}) bounds this confound at $\leq\!1.58$ points, so the
quality gaps partially reflect renderer fidelity differences.

\begin{table*}[!t]
\caption{Cross-Model Comparison on Character Card Generation ($n\!=\!50$).
\textit{Note}: DeepSeek V3 scores are from Figma renders; other models
use preview renders. The RQ3 renderer comparison (Table~\ref{tab:render_comparison})
estimates a Figma advantage of up to $+1.58$ points, implying that
cross-model quality gaps reflect both model capability and renderer
differences.%
\label{tab:model_comparison}}
\centering
\footnotesize
\begin{tabular}{l|ccc|ccccc|c|c}
\toprule
\textbf{Model} & $V$(\%) & $T$(s) & Tokens &
  $S_L$ & $S_C$ & $S_R$ & $S_P$ & $S_A$ & $S_{\text{avg}}$ & \$/call \\
\midrule
DeepSeek V3$^\dagger$       & \textbf{98.0} & 47.0 & 4,534
  & \textbf{8.1$\pm$0.4} & \textbf{8.9$\pm$0.4} & \textbf{7.9$\pm$0.5}
  & \textbf{7.1$\pm$0.6} & \textbf{8.1$\pm$0.5} & \textbf{8.0$\pm$0.3} & 0.008 \\
Gemini 2.0 Flash$^\ddagger$ & 88.0 & \textbf{14.7} & 5,141
  & 6.6$\pm$0.6 & 7.2$\pm$0.8 & 5.6$\pm$0.7 & 6.1$\pm$0.9 & 6.4$\pm$0.6
  & 6.4$\pm$0.6 & \textbf{0.001} \\
GPT-4o-mini$^\ddagger$      & 56.0 & 35.2 & 4,485
  & 6.9$\pm$0.4 & 7.3$\pm$0.8 & 6.0$\pm$0.5 & 6.5$\pm$0.8 & 6.6$\pm$0.5
  & 6.7$\pm$0.4 & 0.012 \\
\bottomrule
\multicolumn{11}{l}{\footnotesize
  $^\dagger$ VLM on Figma renders ($n\!=\!49$).
  $^\ddagger$ VLM on preview renders (Gemini $n\!=\!44$, GPT-4o-mini $n\!=\!28$).}
\end{tabular}
\end{table*}

\begin{figure*}[!t]
\centering
\includegraphics[width=0.9\textwidth]{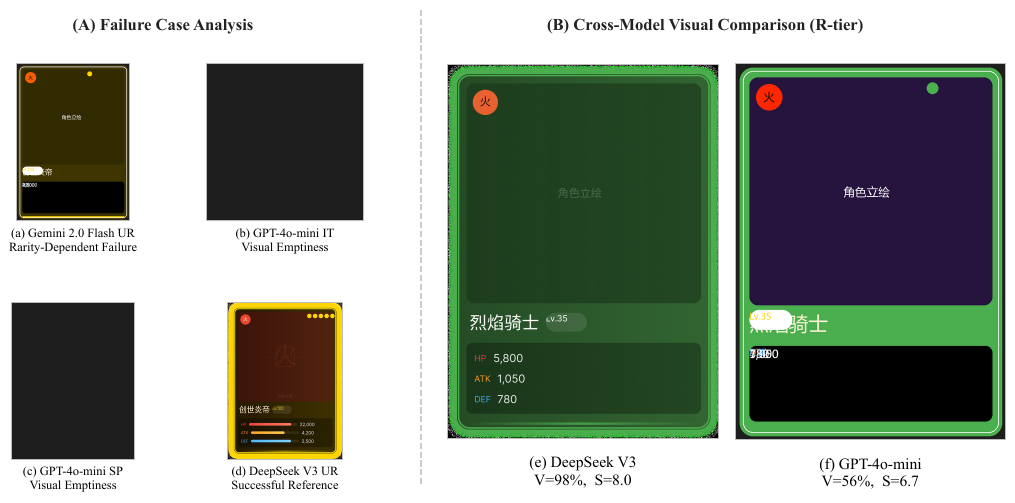}
\caption{Cross-model analysis. \textbf{(A)} Failure cases (left panel):
(a)~Gemini UR-tier degradation ($V\!=\!60\%$); (b--c)~GPT-4o-mini visual
emptiness on Item Thumbnail and Skill Panel; (d)~DeepSeek V3 UR reference.
\textbf{(B)} R-tier comparison (right panel, enlarged for clarity):
DeepSeek V3 (left, $S\!=\!8.0$) produces coherent themed borders,
readable color-coded stats, and a well-structured portrait region;
GPT-4o-mini (right, $S\!=\!6.7$) yields a syntactically valid design
but with low-contrast text, blank portrait region, and element overlap.}
\label{fig:cross_model_combined}
\end{figure*}

Two \textbf{task-structural failure modes} characterize LLM-based
structured UI generation independent of specific model capability levels.
\textbf{Rarity-dependent degradation} manifests with divergent signatures:
Gemini's validity degrades monotonically from 100\% (N-tier) to 60\%
(UR-tier), exposing a \textit{complexity-overload} pattern; GPT-4o-mini
shows an inverse trend (30\% at N), revealing an \textit{over-generation}
pattern on structurally simple inputs. These divergent signatures reflect
each model's distinct structural capacity ceiling rather than general
competence. \textbf{Visual emptiness}---syntactically valid JSON rendering
as blank frames due to zero-area child nodes or indistinguishable
background colors---further establishes that $V\!=\!100\%$ is necessary
but insufficient for design quality, motivating the multi-dimensional
evaluation framework of~\S\ref{sec:experimental_setup}.
Objective structural analysis reinforces these findings: GPT-4o-mini
generates the highest node density (NC\,$=\,26.0\!\pm\!1.6$) and element
completeness (EC\,$=\!0.99$) yet achieves the lowest perceptual scores
($S_{\text{avg}}\!=\!6.7$), confirming that visual harmony outweighs
raw element density. Among all structural metrics, Text Contrast Ratio
is the strongest single predictor of VLM quality for DeepSeek V3
($r\!=\!0.51$, $p\!<\!0.001$, $n\!=\!49$), reflecting the central role
of legibility in game UI appeal.
Fig.~\ref{fig:cross_model_combined} illustrates representative failure cases
and a cross-model visual comparison.

\subsection{RQ2: Ablation and Baseline Study}
\label{sec:ablation}

Table~\ref{tab:ablation} reports ablation results on 50 Character Card
cases (DeepSeek V3). Three findings stand out.

\textbf{Structured prompt design is essential.} The No-Schema baseline
achieves 100\% validity yet VLM quality collapses to 4.8/10
($\Delta\!=\!-3.1$ vs.\ No Few-Shot on matched renderers, $p\!<\!0.001$),
establishing that the schema, domain rules, and color system---not raw
model capability---are the primary quality enablers. Structural metrics
corroborate: Canvas Coverage plummets from 1.96 to 0.05 and Color
Diversity falls 61\%.

\textbf{Post-processing is the critical perceptual driver.} Disabling it
reduces overall quality by 1.4 points ($p\!<\!0.001$, paired Wilcoxon),
with the largest gain in Readability (+2.2), driven by color normalization
correcting text-background contrast deficiencies. Structurally, however,
post-processing leaves the design tree intact: Full Pipeline and No
Post-Process share near-identical structural profiles
(NC\,$=\!23.4$, EC\,$=\!0.93$), confirming it refines within-node
graphical fidelity without altering tree topology---a sub-structural
operation with disproportionate perceptual impact.

\textbf{Few-shot acts as a structural amplifier.} Removing the exemplar
reduces Node Count by 38\%, Canvas Coverage by 39\%, and Color Diversity
by 42\% while exerting negligible impact on VLM perceptual scores
($\Delta\!=\!-0.1$)---demonstrating that the exemplar governs
compositional richness that lies in a \emph{perceptual blind spot}
invisible to VLM evaluation. Counterintuitively, latency increases
(47\,s\,$\to$\,79\,s) despite fewer output tokens (4,534\,$\to$\,3,557),
attributable to increased server-side generation time without a structural
scaffold.

\begin{table}[!tb]
\caption{Ablation and Baseline Study (DeepSeek V3, $n\!=\!50$)%
\label{tab:ablation}}
\centering
\footnotesize
\resizebox{\columnwidth}{!}{%
\setlength{\tabcolsep}{4pt}
\begin{tabular}{l|ccc|cccc}
\toprule
\textbf{Config.} & $V$(\%) & $T$(s) & Tok. &
  $S_{\text{avg}}$ & $S_L$ & $S_R$ & $S_A$ \\
\midrule
Full Pipeline$^\dagger$  & 98.0 & 47 & 4,534
  & \textbf{8.0} & \textbf{8.1} & \textbf{7.9} & \textbf{8.1} \\
No Few-Shot$^\ddagger$   & 96.0 & 79 & 3,557
  & 7.9 & 8.0 & \textbf{7.9} & 7.5 \\
No Post-Proc.$^\ddagger$ & 96.0 & 47 & 4,534
  & 6.5 & 6.7 & 5.7 & 6.5 \\
No-Schema$^\ddagger$     & \textbf{100.0} & 20 & 728
  & 4.8 & 4.9 & 4.7 & 4.8 \\
\midrule
\textit{$\Delta^\ddagger$} & --- & --- & ---
  & \textit{+3.1} & \textit{+3.1} & \textit{+3.2} & \textit{+2.7} \\
\bottomrule
\multicolumn{8}{l}{%
  $^\dagger$Figma renders. $^\ddagger$Preview renders.
  $\Delta^\ddagger$: No-FS minus No-Schema (matched renders).}
\end{tabular}%
}
\end{table}

\subsection{RQ3: Template Generalization and Rendering Fidelity}
\label{sec:rq3}

The pipeline generalizes across all three templates without modification.
Token consumption scales with structural complexity (Item Thumbnail:
1,008; Character Card: 4,534; Skill Panel: 6,165 tokens). Item Thumbnail
achieves 100\% validity at 24.8\,s; Skill Panel shows 90\% validity at
88.7\,s with failures clustering at R/SSR/UR tiers, consistent with the
rarity-complexity effect in RQ1.

\begin{figure}[!tb]
\centering
\includegraphics[width=\columnwidth]{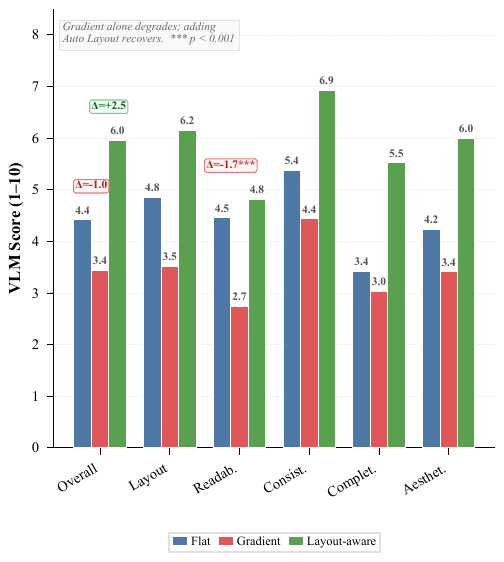}
\caption{Rendering-Evaluation Fidelity: the Gradient renderer
\emph{decreases} VLM score by $-1.00$ ($p\!=\!0.0017$, $d\!=\!-0.81$)
relative to the Flat baseline; the Layout-aware renderer recovers and
surpasses it ($\Delta\!=\!+2.52$, $p\!<\!0.001$, $d\!=\!2.74$).
$n\!=\!27$ Skill Panel designs evaluated across three renderer
configurations with identical Design Spec JSONs.}
\label{fig:two_tier}
\end{figure}

A controlled renderer comparison ($n\!=\!27$ Skill Panels, identical
Design Spec JSONs) reveals the \textbf{Rendering-Evaluation Fidelity
Principle}: adding gradients/shadows \emph{without} layout correction
\emph{degrades} VLM scores from 4.42 to 3.44 ($p\!=\!0.0017$,
$d\!=\!-0.81$). The Layout-aware renderer---which resolves child
coordinate conflicts via Auto Layout simulation \emph{before}
rasterization---recovers to 5.96 ($\Delta\!=\!+2.52$,
$p\!<\!0.001$, $d\!=\!2.74$). Table~\ref{tab:render_comparison} reports
the full per-dimension breakdown. Gradient shading amplifies the visual
salience of overlapping elements (Readability drop: $-1.73$,
$p\!<\!0.001$, $d\!=\!-1.38$), converting latent layout defects into
highly visible flaws that the VLM penalizes sharply.
Fig.~\ref{fig:two_tier} visualizes this two-tier bottleneck.

\begin{table}[!tb]
\caption{Controlled Renderer Comparison: Same 27 Skill Panel Designs
Evaluated Across Three Configurations (paired Wilcoxon signed-rank)%
\label{tab:render_comparison}}
\centering
\footnotesize
\resizebox{\columnwidth}{!}{%
\setlength{\tabcolsep}{3pt}
\begin{tabular}{l|ccc|c|cc}
\toprule
\textbf{Renderer} & $S_{\text{avg}}$ & $S_R$ & $S_C$ &
  $\Delta$ vs.\ prev & $p$ & $d$ \\
\midrule
Flat (solid colors only)                 & 4.42{\scriptsize$\pm$1.01} & 4.46 & 5.38 & ---         & ---      & ---   \\
Gradient (+grad./shadows, no layout)     & 3.44{\scriptsize$\pm$0.63} & 2.74 & 4.44 & $-$1.00$^a$ & 0.0017   & $-$0.81 \\
Layout-aware (+Auto Layout simulation)   & \textbf{5.96{\scriptsize$\pm$0.84}} & \textbf{4.82} & \textbf{6.93} & $+$2.52$^b$ & $<$0.001 & \textbf{2.74} \\
\midrule
\textit{Net Flat$\to$Layout-aware}       & --- & --- & --- & $+$1.58$^c$ & 0.0001   & 1.16  \\
\bottomrule
\multicolumn{7}{l}{\footnotesize $^a$ Gradient amplifies overlap salience ($S_R$: $-$1.73, $p\!<\!0.001$, $d\!=\!-1.38$).} \\
\multicolumn{7}{l}{\footnotesize $^b$ Auto Layout simulation eliminates overlaps before rasterization.} \\
\multicolumn{7}{l}{\footnotesize $^c$ Net $S_R$ not significant ($+$0.38, $p\!=\!0.18$); Tier-1 harm partially offsets recovery.}
\end{tabular}%
}
\end{table}

\subsection{RQ4: Agentic Refinement via Reflection Controller}
\label{sec:rq4}

Table~\ref{tab:reflection} reports RC results on 28 valid Character Card
cases (DeepSeek V3, Gradient renderer, $K\!\leq\!2$, $\theta\!=\!7.5$).
The Gradient renderer is used here rather than Layout-aware, so that RC
gains are attributable to the iterative repair loop rather than
renderer-side overlap resolution---thereby isolating the RC's intrinsic
correction capacity.
The RC achieves mean $\Delta\!=\!+0.96$ ($W\!=\!0$, $p\!<\!0.001$,
$d\!=\!1.30$, 95\%\,CI $[+0.68,\,+1.25]$). Best-result tracking yields 100\% regression-free outcomes (all 28
cases at or above initial score); only 7.7\% experience a transient
Iter-1 score dip. The best single-case gain is
CC-020 ($5.0\!\to\!8.0$, $\Delta\!=\!+3.0$). Gains are consistent across
rarity tiers (N: $+1.10$, R: $+1.11$, SR: $+0.67$).

\begin{table}[!tb]
\caption{Reflection Controller Results by Rarity Tier (DeepSeek V3,
Gradient renderer, $K\!\leq\!2$, $\theta\!=\!7.5$;
Wilcoxon: $W\!=\!0$, $p\!<\!0.001$, $d\!=\!1.30$).%
\label{tab:reflection}}
\centering
\footnotesize
\resizebox{\columnwidth}{!}{%
\setlength{\tabcolsep}{4pt}
\begin{tabular}{l|ccc|c}
\toprule
\textbf{Metric} & \textbf{N} ($n\!=\!10$) & \textbf{R} ($n\!=\!9$) &
  \textbf{SR} ($n\!=\!9$) & \textbf{All} ($n\!=\!28$) \\
\midrule
Init $S_{\text{avg}}$    & 5.30{\scriptsize$\pm$0.48} & 5.56{\scriptsize$\pm$0.78}
  & 7.00{\scriptsize$\pm$0.50} & 5.93{\scriptsize$\pm$0.98} \\
Final $S_{\text{avg}}$   & 6.40{\scriptsize$\pm$0.70} & 6.67{\scriptsize$\pm$1.00}
  & 7.67{\scriptsize$\pm$0.50} & \textbf{6.89{\scriptsize$\pm$0.92}} \\
Avg $\Delta$             & +1.10 & +1.11 & +0.67 & \textbf{+0.96} \\
Improved / Same          & 9/1 & 7/2 & 5/4 & 21/7 \\
Regression-free          & 10/10 & 9/9 & 9/9 & \textbf{28/28 (100\%)} \\
Final $\geq\!7.5$        & 0/10 & 2/9 & 6/9 & 8/28 (28.6\%) \\
\bottomrule
\multicolumn{5}{l}{\footnotesize
  \textit{Note:} SSR/UR excluded (init~$\geq\!\theta$); headroom at
  init$\!=\!7.66$ gives $\Delta\!=\!+0.07$ (Quality Ceiling Effect).}
\end{tabular}%
}
\end{table}

\begin{figure}[!tb]
\centering
\includegraphics[width=\columnwidth]{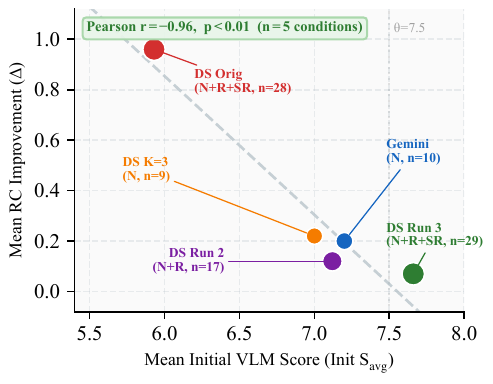}
\caption{Quality Ceiling Effect: Pearson $r\!=\!-0.96$ ($p\!<\!0.01$)
between mean initial VLM score and RC improvement ($\Delta$) across five
independent conditions ($n_{\text{total}}\!=\!93$; DeepSeek~V3 and
GPT-4o-mini; five tier mixes). RC gain is universally bounded by
headroom below $\theta$, not by model capability or rarity complexity.}
\label{fig:ceiling}
\end{figure}

\begin{figure}[!tb]
\centering
\includegraphics[width=\columnwidth]{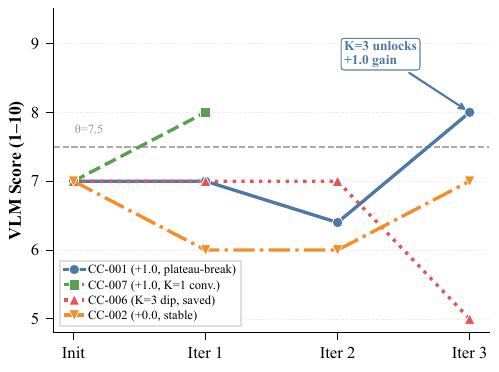}
\caption{$K\!\leq\!3$ N-tier ablation: RC score trajectories for
representative cases. CC-001 exhibits a plateau-breaking pattern
($7.0\!\to\!7.0\!\to\!6.4\!\to\!8.0$) inaccessible at $K\!\leq\!2$;
CC-007 converges at $K\!=\!1$; best-result tracking prevents CC-006's
$K\!=\!3$ regression. Dashed line: $\theta\!=\!7.5$.}
\label{fig:score_trajectory}
\end{figure}

\textbf{Quality Ceiling Effect.} Compiling five independent RC conditions spanning DeepSeek~V3 and
GPT-4o-mini configurations, five tier mixes, and three iteration
budgets ($n_{\text{total}}\!=\!93$ valid cases), RC improvement exhibits a
near-perfect negative correlation with mean initial quality: Pearson
$r\!=\!-0.96$ ($p\!<\!0.01$, $t\!=\!-5.9$, $\text{df}\!=\!3$), spanning
initial scores from 5.93 ($\Delta\!=\!+0.96$) to 7.66 ($\Delta\!=\!+0.07$).
Given the five-condition aggregate (df\,=\,3), this correlation is
exploratory evidence for the headroom hypothesis rather than a precisely
fitted universal law; replication with additional generator
configurations is warranted.
This documents \emph{critique saturation}: as designs approach $\theta$,
VLM critiques shift to increasingly marginal issues, diminishing repair
precision. Skip rates (initial score $\geq\!\theta$) scale monotonically
with initial quality from 3.6\% to 65.5\%, and all 93 cases remain
regression-free. The renderer comparison---Flat ($\Delta\!=\!0.00$, dip
rate 76\%) $\to$ Gradient ($\Delta\!=\!+0.96$, dip rate 7.7\%)---indicates
evaluation fidelity as a primary limiting factor for RC effectiveness.
Fig.~\ref{fig:ceiling} visualizes the Quality Ceiling Effect.
Fig.~\ref{fig:score_trajectory} shows trajectories from a supplementary
$K\!\leq\!3$ ablation on the 10 N-tier cases (DeepSeek V3, Gradient
renderer): CC-001 exhibits a plateau-breaking pattern
($7.0\!\to\!7.0\!\to\!6.4\!\to\!8.0$) inaccessible at $K\!\leq\!2$;
best-result tracking prevents CC-006's $K\!=\!3$ regression.

\section{Discussion}

\textbf{Design Principles.}
The ablation establishes an actionable hierarchy for LLM-driven UI
generation: \emph{structured prompt design} is the foundational quality
enabler; \emph{post-processing} is the primary perceptual driver
(+1.4 VLM, +2.2 Readability; Wilcoxon, No-FS vs.\ No-Post-Proc on
matched preview renders); and the \emph{few-shot exemplar} functions as a
structural amplifier (38\% Node Count gain) whose contribution is
invisible to VLM perceptual evaluation but critical for compositional
richness. All three components are non-negotiable in production deployment.

\textbf{Rendering-Evaluation Fidelity Principle.}
Formally, let $\hat{q}_t = \text{VLM}(\text{render}(d_t))$ denote the
observable quality signal for design $d_t$ with true quality $q_t$. Under
the Gradient renderer, $\hat{q}_t < q_t$: gradient shading amplifies the
salience of overlapping elements concealed by flat colors, inverting the
VLM reward signal. Under the Layout-aware renderer, $\hat{q}_t \approx q_t$,
since Auto Layout simulation resolves overlaps prior to graphical
enhancement. Consequently, synchronized improvements across both rendering
tiers---graphical fidelity and spatial correctness---are necessary for
reliable agentic refinement. The Quality Ceiling Effect ($r\!=\!-0.96$)
quantifies reward saturation in the visual domain, drawing a
direct parallel to test-time compute scaling laws~\cite{snell2024scaling}.
These principles apply broadly to any LLM agent that synthesizes
structured visual specifications and refines them based on rendered
perceptual feedback.

\textbf{Practical RC Deployment.}
The Quality Ceiling Effect translates directly into a tier-aware compute
allocation policy: (i)~if init$\!<\!6.0$, apply $K\!\leq\!2$ RC cycles
(expected $\Delta\!>\!+0.9$); (ii)~if $6.0\!\leq\!$init$\!<\!7.5$,
$K\!\leq\!2$ is recommended ($\Delta\!\approx\!+0.5$--$0.7$);
(iii)~if init$\!\geq\!7.5$, RC should be skipped ($\Delta\!\approx\!0$);
and (iv)~upon detecting a late-iteration score dip within $K\!\leq\!2$
(e.g., $s_{k=2}\!<\!s^*$), extending to $K\!\leq\!3$ may exploit a
plateau-breaking event (as in CC-001,
$7.0\!\to\!7.0\!\to\!6.4\!\to\!8.0$, inaccessible at $K\!\leq\!2$).
The threshold $\theta\!=\!7.5$ performs consistently across the
conditions evaluated here; practitioners should calibrate it to
approximately the 80th percentile of their generator's output distribution.

\textbf{Limitations.}
The evaluation covers RPG gacha UIs exclusively; FPS HUDs and RTS
minimaps remain unvalidated. The current schema supports four primitive
node types, whereas production demands vector paths and image fills.
Portrait regions are rendered as placeholders, and animation or
interactive states are not modeled. Cross-template comparisons involve
heterogeneous rendering environments; this confound is bounded at
$\leq\!0.1$ points by restricting $\Delta$ comparisons to
renderer-consistent conditions. The cross-model comparison (RQ1) reflects three representative API
configurations spanning the quality--cost--speed design space; the
identified failure modes---rarity-dependent degradation and visual
emptiness---are posited as task-structural properties of LLM-based
structured generation, expected to persist across model generations, while
specific validity rates and quality scores will shift with newer generators.
The framework's model-agnostic design and released 110-case evaluation
suite support direct replication with current SoTA models.
Beyond model coverage, the primary limitation of the current work
is the absence of human expert evaluation.
All quality assessments rely on GPT-4o as an automated VLM critic;
while ICC(2,1)$\,=\,0.555$ confirms adequate within-evaluator
reliability for comparative analyses, external validity---the
correlation between VLM scores and professional designer
judgment---remains unestablished. Recent work on LLM-as-Judge has
demonstrated moderate-to-strong agreement with human annotators in
adjacent domains~\cite{zheng2023judging}, and our per-design
stability (mean MAD$\,=\,0.15$ across three evaluation runs) suggests
systematic rather than random scoring. Nevertheless, a formal human
expert study is a necessary next step for deployment validation.

\section{Conclusion}

We present GameUIAgent, a neuro-symbolic framework bridging natural
language descriptions and professional game UI design through a
\textit{Design Spec JSON} intermediate representation. By combining LLM
generation with an independent VLM-driven Reflection Controller, the
system delivers editable Figma outputs with a formal guarantee of
non-regressive quality improvement across 110 evaluated cases.

A cross-model failure taxonomy establishes \textit{rarity-dependent
degradation} and \textit{visual emptiness} as task-structural failure
modes persistent across LLM families, with structural analysis revealing
a \textit{perceptual blind spot} where few-shot scaffolding governs
compositional richness ($+38\%$ NC, $+42\%$ CD) invisible to VLM
perceptual scoring. Two empirical findings further establish foundational
principles for visual agentic systems. The \textbf{Quality Ceiling Effect}
($r\!=\!-0.96$) reveals that iterative improvement is bounded by
evaluator headroom---where \emph{evaluation capability}, not generation
power, becomes the limiting factor. The \textbf{Rendering-Evaluation
Fidelity Principle} warns that incomplete rendering enhancements
paradoxically invert reward signals by amplifying structural defects. As
AI tools evolve from single-shot generation toward autonomous iterative
refinement in game production, rigorous alignment among structural
constraints, rendering fidelity, and evaluation mechanisms will be
paramount. The framework's model-agnostic design and released 110-case
evaluation suite enable direct replication as more capable LLMs emerge,
extending these findings beyond the three model configurations evaluated
here.

\FloatBarrier


\end{document}